\documentclass{article} 
\usepackage{iclr2024_conference,times}

\usepackage{amsmath,amsfonts,bm}

\def\eqref#1{equation~\ref{#1}}

\def\1{\bm{1}}

\DeclareMathAlphabet{\mathsfit}{\encodingdefault}{\sfdefault}{m}{sl}
\SetMathAlphabet{\mathsfit}{bold}{\encodingdefault}{\sfdefault}{bx}{n}

\usepackage{hyperref}
\usepackage{url}

\usepackage{color,colortbl}
\definecolor{color1}{HTML}{006EB8}
\definecolor{color2}{HTML}{009B55}

\usepackage[utf8]{inputenc} 
\usepackage[T1]{fontenc}    
\hypersetup{colorlinks=true,urlcolor=color2,linkcolor=color2,citecolor=color2}
\usepackage{booktabs}       
\usepackage{amsfonts}       
\usepackage{nicefrac}       
\usepackage{microtype}      
\usepackage{xcolor}         
\usepackage[pdftex]{graphicx}
\usepackage{wrapfig}
\usepackage{amsmath}

\newcommand{\algo}[1]{$\mathbb{D}^2$ \textsc{Pruning}}

\title{$\mathbb{D}^2$ Pruning: Message Passing for Balancing \\
Diversity \& Difficulty in Data Pruning}

\author{Adyasha Maharana \;\;\; Prateek Yadav \;\;\; Mohit Bansal \\
Department of Computer Science\\
University of North Carolina Chapel Hill\\
Chapel Hill, NC 27510, USA \\
\texttt{\{adyasha,praty,mbansal\}@cs.unc.edu} \\
}

\iclrfinalcopy 
\begin{document}

\maketitle

\begin{abstract}
In recent years, data quality has emerged as an important factor for training massive models. Analytical theories suggest that higher-quality data can lead to lower test errors in models trained on a fixed data budget. Moreover, a model can be trained on a lower compute budget without compromising performance if a dataset can be stripped of its redundancies. Coreset selection (or data pruning) seeks to select a subset of the training data so as to maximize the performance of models trained on this subset, also referred to as coreset. There are two dominant approaches: (1) geometry-based data selection for maximizing \textit{data diversity} in the coreset, and (2) functions that assign \textit{difficulty scores} to samples based on training dynamics. Optimizing for data diversity leads to a coreset that is biased towards easier samples, whereas, selection by difficulty ranking omits easy samples that are necessary for the training of deep learning models. This demonstrates that data diversity and importance scores are two complementary factors that need to be jointly considered during coreset selection. In this work, we represent a dataset as an undirected graph and propose a novel pruning algorithm, \algo{}, that uses forward and reverse message passing over this dataset graph for coreset selection. \algo{} updates the difficulty scores of each example by incorporating the difficulty of its neighboring examples in the dataset graph. Then, these updated difficulty scores direct a graph-based sampling method to select a coreset that encapsulates both diverse and difficult regions of the dataset space. We evaluate supervised and self-supervised versions of our method on various vision and language datasets. Results show that \algo{} improves coreset selection over previous state-of-the-art methods for up to 70\% pruning rates. Additionally, we find that using \algo{} for filtering large multimodal datasets leads to increased diversity in the dataset and improved generalization of pretrained models. Our work shows that \algo{} is a versatile framework for understanding and processing datasets.\footnote{Our code is available at \url{https://github.com/adymaharana/d2pruning}}
\end{abstract}

\section{Introduction}
Deep learning models are evolving into massive architectures with trillions of learnable parameters requiring enormous training datasets for optimal performance. Empirical experiments demonstrate that the test error in such models falls off as a power law with model size as well as training dataset size \citep{kaplan2020scaling}. Recently, \citet{sorscher2022beyond} developed an analytical theory that shows that the power law association of test error with data size can be demoted to exponential scaling if one has access to a high-quality \textit{data pruning} metric for careful data selection. This has the implication that for a fixed data budget, high-quality training data can yield lower test loss in deep learning models. \textit{Coreset selection} \footnote{We use the terms coreset selection and data pruning interchangeably throughout the paper.} \citep{mirzasoleiman2020coresets, guo2022deepcore} is a similar line of work that aims to select a subset (coreset) of the most informative samples $\mathcal{S}$ from a large training dataset $\mathcal{T}$ without significantly compromising the performance of the model. Existing coreset selection methods \citep{tonevaempirical, killamsetty2021glister, yang2022dataset, sorscher2022beyond} demonstrate promising performance on many vision datasets for one-shot coreset selection. However, significant progress remains to be made on the selection of better coresets, especially using self-supervised approaches. Moreover, there is a lack of systematic evaluation of these methods on NLP datasets.

\begin{figure}[t]
    \centering
    \includegraphics[width=0.98\linewidth]{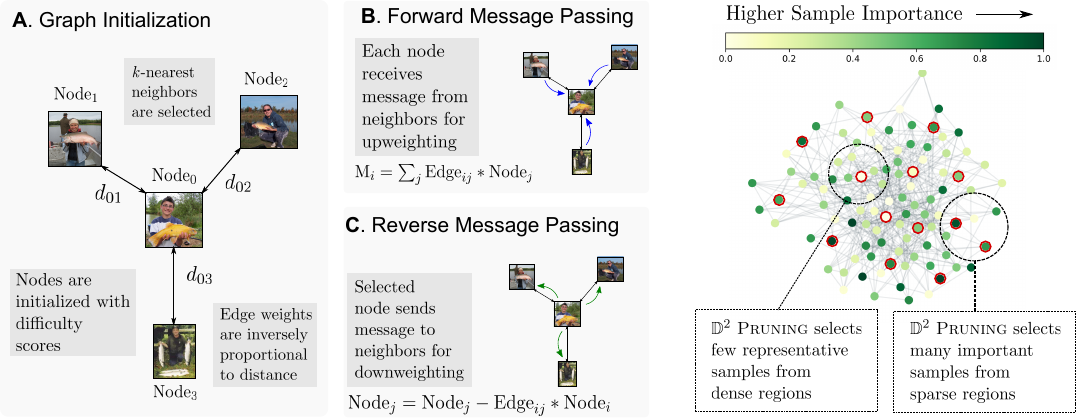}
    \caption{Overview of \algo{}. (left) Our proposed algorithm contains three steps: (a) Initialization of graph $\mathcal{G}$ using difficulty scores and edge weights based on embedding distance, (b) message passing between connected nodes to propagate difficulty scores of neighboring samples, and (c) data selection and reverse message passing to avoid sampling from the same neighborhood. (right) \algo{} selects a balanced subset of samples (red) from sparse and dense regions.}
    \label{fig:method}
    \vspace{-10pt}
\end{figure}

Real-world data distributions comprise high-density as well as low-density regions. \citet{yu2020learning, chan2022redunet} claim that maximizing the variance of intra-class embeddings results in robust representations. To this end, geometry-based coreset selection methods \citep{seneractive, chen2010super} operate under the assumption that samples located close to each other provide redundant information, and try to remove those data points by selecting the samples most distant from $k$-means cluster centers \citep{sorscher2022beyond} or at a median distance from the class center \citep{xia2023moderate}, in order to maximize diversity in the coreset. On the other hand, uncertainty-based methods \citep{colemanselection} and error or loss-based methods \citep{tonevaempirical, paul2021deep} propose a score-based function to estimate the difficulty of each sample in the training dataset from the model's training dynamics and retain the most difficult samples. 
However, the distribution of difficulty scores for the original data is highly skewed and contain way more low-difficulty (or easy) samples \citep{swayamdipta2020dataset}, as we show in Figure~\ref{fig:sampling}(a). As low-difficulty samples predominantly arise in densely populated regions \citep{sorscher2022beyond}, incorporating some of these well-connected, low-difficulty samples into the coreset guarantees adequate representation of these dense areas within the coreset \citep{zheng2022coverage}. At the same time, selecting high-difficulty samples with higher connectivity increases the information content of the coreset. Evidently, \textit{example difficulty} and \textit{data diversity} are two crucial factors for selecting effective coresets, yet, there has been little work towards combining the two factors into a unifying framework for coreset selection.

To unify these two factors, we propose the \algo{} method, where we represent the dataset $\mathcal{S}$ as an undirected graph $\mathcal{G}$ and design a message-passing algorithm that unifies the difficulty scores and the underlying spatial distribution of the dataset to select a coreset with balanced difficulty and diversity. \algo{} consists of three simple steps: 
(A) \textbf{Graph Initialization:} First, we create a \textit{graph}, $\mathcal{G}$, where each node is an example from the dataset $\mathcal{S}$ and is connected to its $k$-closest neighbors based on a notion of distance in the embedding space (see Fig.~\ref{fig:method}(A)). Each node has a feature value that represents the \textit{difficulty score} of the example. This graph can be used to understand the connectivity of each sample with respect to the rest of the dataset \citep{ebert2012ralf}.
(2) \textbf{Forward Message Passing:} Next, we perform message passing \citep{gasteigerdirectional,yadav19lcn} over the dataset graph to update the difficulty scores of all examples by taking into account the distance and difficulty of its neighboring examples in the graph (see Fig.~\ref{fig:method}(B)). Specifically, each node collects a message from all of its neighbors (where the message is their difficulty scores scaled by their distance) and uses these messages to update its own difficulty score.
(3) \textbf{Coreset Selection \& Reverse Message Passing:} Finally, we use these updated scores to iteratively select a balanced subset of samples from high-density low-difficulty regions and low-density high-difficulty regions. At each step of selection, the neighbors of the selected sample are down-weighted via reverse message-passing to promote diversity in the coreset (see Fig.~\ref{fig:method}(C)). Our design ensures that highly connected nodes of low difficulty are on equal footing with sparsely connected nodes of high difficulty during selection.

We refer to this diversity-difficulty ($\mathbb{D}^2$) approach of coreset selection using message-passing as \algo{} and evaluate this pruning method on multiple image classification and natural language processing (NLP) datasets. We find that \algo{} outperforms state-of-art methods for coreset selection at low-to-medium pruning rates. Our analysis shows that \algo{} selects a coreset with a higher distribution of difficult samples for low pruning rates and with equitable distribution over easy and difficult samples for medium-to-high pruning rates. Further, we adapt \algo{} for self-supervised and unsupervised data selection approaches and show improvements over existing methods for self-supervised coreset selection and data filtering respectively. Importantly, the message-passing framework for coreset selection opens up possibilities for exploring different message schemes, possibly incorporating factors other than data diversity and difficulty, in an easy plug-and-play framework. In summary, our contributions are: 

\begin{itemize}
    \item We propose \algo{}, a one-shot coreset selection algorithm that represents datasets as undirected graphs and uses message-passing to combine the influence of two important factors, \textit{example difficulty} and \textit{data diversity}, for data selection.
    \item We evaluate our method on several image classification and NLP benchmarks and show state-of-the-art results for low-to-medium pruning rates for supervised as well as self-supervised approaches. To the best of our knowledge, we are the first to perform a systematic evaluation of coreset selection methods on NLP datasets.
    \item We show that \algo{} selects diverse data pools when filtering massive multimodal datasets, which improves the generalization of pretrained multimodal models.
\end{itemize}

\section{Preliminaries}
\label{sec:prelim}

In this section, we describe one-shot coreset selection and discuss the motivation behind our work.

\subsection{One-Shot Coreset Selection}
\label{sec:oneshot}
Consider a training dataset $S$ containing $N$ examples $\{(x_i, y_i)\}^{N}_{i=1}$ drawn i.i.d. from an underlying distribution $P$. One-shot coreset selection refers to the selection of a subset $S'$ of the data at a given pruning rate $\alpha$ such that the loss of the model $\theta$ trained on $S'$ using loss function $L$ is minimized on an evaluation set drawn from $P$. This results in the optimization problem as follows:
\begin{equation}
    \min_{S' \subset S: \frac{|S'|}{|S|} \le (1 - \alpha)} E_{x,y \sim P} [L(x, y;\theta)]
\end{equation}

\subsection{Desiderata of Coreset}
\label{sec:desiderata}
Coresets are representative subsets of larger datasets and aim to preserve the performance achieved by training on the full dataset. Prior works on understanding training dynamics point towards two important factors for ensuring the same i.e. example difficulty and data diversity.

\paragraph{Example difficulty.} Multiple works have sought to define example difficulty in order to understand how deep neural networks process data. Statistical metrics like consistency score \citep{jiang2021characterizing} measure the probability of predicting the correct label of an instance when it is left out of the training dataset. \citet{sorscher2022beyond} provide theoretical justification for retaining the hardest examples when pruning large datasets for a perceptron learning setting. \citet{swayamdipta2020dataset} show that examples that have a high degree of variance in the model's predictions during training have the largest impact on the model's overall performance. Accordingly, coreset selection methods based on difficulty score functions prioritize the selection of difficult examples for coresets \citep{guo2022deepcore}. However, it has been shown that deep learning models learn easy data and simple functions earlier in training \citep{jiang2021characterizing, tonevaempirical, baldock2021deep} and easy examples ease the optimization of deep learning networks in the high-dimensional data manifold. Moreover, \citet{zheng2022coverage} demonstrate that it is necessary to include easy examples to ensure coverage in high-density areas of the data distribution, which leads to the next factor of consideration i.e. data diversity.

\begin{figure}
    \centering
    \includegraphics[width=0.95\linewidth]{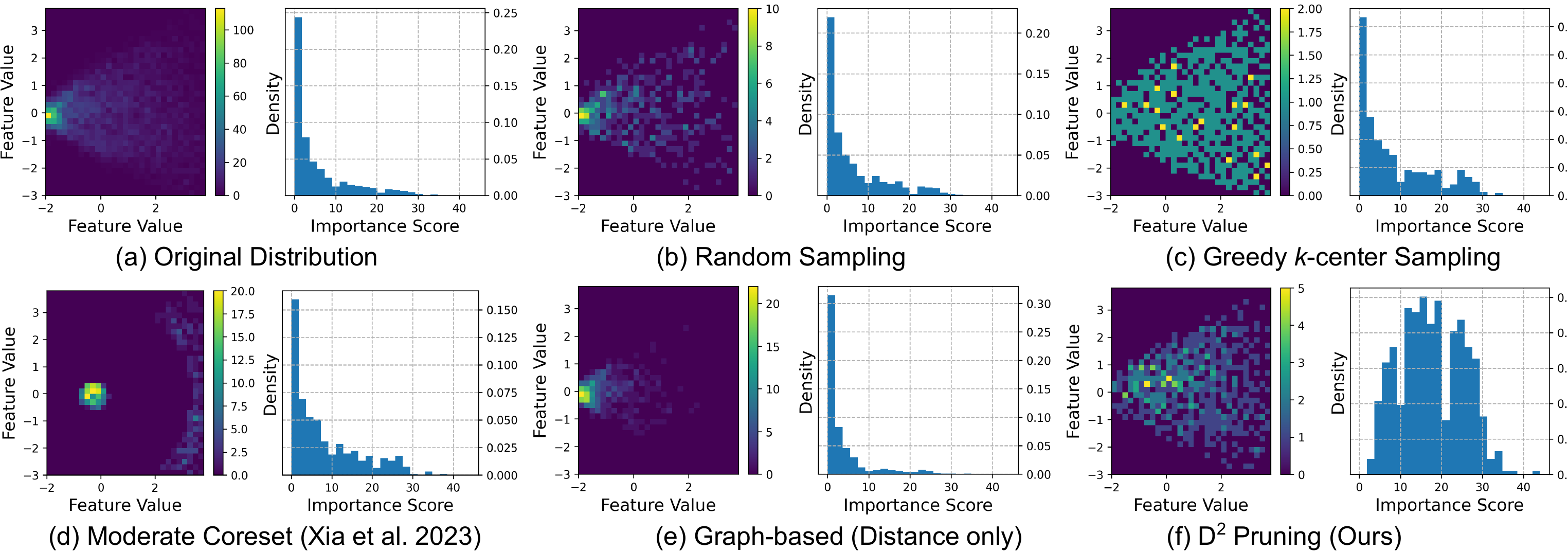}
    \caption{Sampling Methods. Demonstration of data distribution (left) and importance scores (right) in (a) a single class in the CIFAR10 dataset, and coresets selected under 90\% pruning rate via (b) random sampling, (c) greedy $k$-center selection that maximizes data diversity, (d) moderate coreset \cite{xia2023moderate} (e) graph-based density sampling using embedding distance \citep{ebert2012ralf} and (f) our method, \algo{}, designed to balance data diversity and difficulty during coreset selection. Embeddings are extracted from a ResNet18 model trained on CIFAR10.}
    \label{fig:sampling}
    \vspace{-5pt}
\end{figure}

\paragraph{Data diversity.} Representation structure has been explored in several works as the key to the generalization of deep learning models; variance in representations for each class should be as large as possible while also being uncorrelated from other classes \citep{xia2023moderate}. The diversity of a dataset can be captured in many ways such as coding rate \citep{yu2020learning, chan2022redunet}, max dispersion or convex hull volume \citep{yu2022can} and coverage \citep{seneractive, zheng2022coverage}. A set $S^\prime$ is a $r$-cover of another set $S$, when a set of $r$-radius balls centered at each element in $S^\prime$ covers the entire $S$. The radius $r$ can be used as a metric to measure coverage of $S^\prime$ on $S$ \citep{seneractive}. \citet{zheng2022coverage} introduce the metric AUC$_{pr}$ (Area under coverage), which is computed against test set $D_{test}$ i.e. AUC$_{pr}$ $(S) = E_{x\in D_{test}} [min_{x^{\prime}\in S} d(x^{\prime}, x)]$ and theoretically show that it is important to minimize the AUC$_{pr}$ for better generalization. Difficult samples tend to be rarer samples found in the low-density areas of the data distribution whereas easy samples tend to lie in high-density areas. An effective coreset should contain sufficient samples from both areas to ensure maximum coverage. However, optimizing for diversity only leads to coresets with a skewed distribution over example difficulty. As we show in Fig.~\ref{fig:sampling}(c), $k$-center selection minimizes the distance of samples in $S$ from $S'$ and has high coverage of the underlying data distribution. But, the selected coreset contains a disproportionate number of easy samples, rendering it ineffective.

\textit{Example difficulty} and \textit{diversity} are two complementary factors that make an effective coreset. Hence, coreset selection methods need to unify the influence of these factors in a constructive manner. To this end, we represent the dataset $S$ as a graph and introduce a novel message-passing algorithm~\citep{vashishth2019wordgcn,vashishth19confgcn}, \algo{}, that accounts for both factors when selecting samples for coreset.

\section{\algo{}: Message passing for coreset selection}
\label{sec:our_method}
Consider a dataset $\mathcal{S}$, where each sample $s$ is represented in an embedding space, i.e., $s\in\mathbf{R}^{d}$. We seek to select a coreset $S'$ consisting of a subset of the samples in $\mathcal{S}$ as outlined in Sec.~\ref{sec:oneshot}. Moreover, our goal is to combine the influence of embedding distance and difficulty scores when selecting samples for coreset (see Sec.~\ref{sec:desiderata}). This setting naturally lends itself to a representation using undirected graph $\mathcal{G}$, where each sample is represented as a node with node-feature $x_{i}$, and edge weights $e_{ij}$ to indicate its connectivity with other samples in the embedding space (see Fig.~\ref{fig:method}(a)). We use message-passing to `inform' a sample about (a) its proximity to adjacent samples in an embedding space, and (b) the difficulty scores of its neighbors. First, we briefly discuss message passing for graphs, and then we discuss our proposed algorithm, \algo{}.

\subsection{Message Passing}
Message passing \citep{hamilton2017inductive} is a widely-used operation performed on graphs to propagate information from a node's neighbors to itself and update the state of the node based on the newly acquired information. For instance, \citet{gilmer2017neural, gasteigerdirectional} use message-passing to encode molecular structures for chemical prediction. The message-passing phase is defined in terms of a message function $M$ and a node update function $U$. In the message passing phase, a given node $i$ receives messages from each of its neighbors and aggregates them as follows to update its own feature value as,
\begin{align}
    m_{i} &= \sum_{j\in \mathcal{N}(i)} m_{ij}~;~~~ \text{where}~~ m_{ij} = M (x_{j}, e_{i,j}) \\
    x_{i} &= U(x_{i}, m_{i})
\end{align}
where $\mathcal{N}(i)$ denotes the neighbors of node ${i}$ in graph $\mathcal{G}$. $U$ is an aggregation function that accounts for the messages received from all neighbors, as well as the node's own feature.

\subsection{\algo{}}
\algo{} consists of 3 stages i.e., (a) Graph initialization, (b) forward message passing, and (c) data selection via reverse message passing.

\paragraph{Graph initialization.} We create a single, sparse graph for the dataset $S$ where each sample in $S$ is represented by a node $i$ in the graph. In order to account for example difficulty during coreset selection, we initialize the node feature as the difficulty score of the sample based on training dynamics of the model $\theta$ trained on $\mathcal{S}$, i.e., $x_{i} = f_{\theta}(s_{i})$, where $f(.)$ is the scoring function. In practice, the scoring function can be one of the many metrics used to measure difficulty such as forgetting \citep{tonevaempirical}, consistency score \citep{jiang2021characterizing}, and self-supervised metrics like prototypicality \citep{sorscher2022beyond} etc. Next, we collect the $k$ nearest neighboring samples for every sample in the dataset. Within the graph, the connecting edges between each node $i$ and its $k$ nearest neighbors are initialized with a non-zero edge weight $e_{i,j}$, where node $j$ is one of the $k$ nearest neighbors (see Fig.~\ref{fig:method}(a)). All other edge weights are set to zero, leading to a sparse graphical representation of the entire dataset $S$. The edge weight $e_{i,j}$ represents the proximity of the two nodes $i, j$ using the RBF kernel of the distance $d(i,j)$. We use the Euclidean distance as the distance function i.e., $d(i,j) = ||v_{i}-v{j}||$ where $v_{i}$ is the embedding vector for sample $i$.

\paragraph{Forward message passing.} In this step, each node $i$ in the graph receives information about its neighborhood via a single step of message propagation. Every connected node $j$ sends a message $M$ to node $i$ about its importance score which is scaled by the edge weight as,
\begin{equation}
    M(x_{j}, e_{ij}) = e_{i,j} * x_j~;~~~ \text{where}~~e_{i,j} = \exp{(-\gamma_{f}*d(i,j)^2)}
\end{equation}
The intuition behind this definition is that samples that are farther away from the node but are of higher difficulty should be weighted similarly to samples that are closer to the node and have lower difficulty. This promotes diversity in the coreset by ensuring representation from all regions of the data distribution. Finally, the receiving node $i$ aggregates all of the messages received from its neighboring nodes and updates its own feature value as,
\begin{equation}
    U_{f}(x_{i}, m_{i}) = x_{i} + \sum_{j\in \mathcal{N}(i)} M (x_j,e_{i,j})
\end{equation}
This reinforces the importance of dense regions comprising easy samples or sparse regions comprising difficult samples. Therefore, in this way, we start with a graph $\mathcal{G}$ where connectivity is based on the distance between two samples in the embedding space and convert it into a graph based on distance as well as difficulty scores via message passing.

\paragraph{Data selection via reverse message passing.} In the final step, samples in $\mathcal{S}$ are ranked according to their corresponding \textit{updated} node feature values in $\mathcal{G}$. Iteratively, the highest ranking sample $s_{k} = \arg \max_{i\in\mathcal{S}}x_{i}$ is selected \citep{ebert2012ralf}, and its neighboring nodes are down-weighted to maximize the diversity of the coreset. However, since the distance between two nodes is a representation of their semantic similarity, neighboring nodes that are farther away from the selected node must be down-weighted relatively less than those that are closer. We implement this via reverse message passing, where the neighboring nodes receive a weighted message from the selected node and use it to update their feature value as,
\begin{equation}
    x_{j} = x_{j} - e_{k,j} * x_{k}, ~~\forall j\in \mathcal{N}(k)~;~~~ \text{where}~~e_{k,j} = \exp{(-\gamma_{r}*d(k,j)^2)},
\end{equation}
where a lower value of $\gamma_{r}$ causes larger updates in connected nodes and vice-versa. With these steps, \algo{} selects a coreset that contains samples from all regions of the data distribution and are more uniformly distributed over the range of difficulty scores (see Fig.~\ref{fig:sampling}(f)). In the following sections, we use this framework for supervised, self-supervised approaches to coreset selection and as a filtering strategy for massive unlabelled datasets.

\section{Experimental Setup}
\label{sec:setup}

\paragraph{Tasks, Models \& Datasets.} We evaluate \algo{} on three vision datasets i.e., \textbf{CIFAR10}, \textbf{CIFAR100} \citep{krizhevsky2009learning} and \textbf{Imagenet-1K} \citep{deng2009imagenet}, and two NLP datasets i.e., a subset (2k train examples) of \textbf{ImDB} reviews for sentiment analysis, and the Adversarial NLI (\textbf{ANLI}) dataset \citep{nie2020adversarial} for natural language inference. To the best of our knowledge, we are the first to perform a systematic evaluation of coreset selection methods on NLP datasets. We evaluate unsupervsied \algo{} on the DataComp (small) dataset \citep{gadre2023datacomp}. We use ResNet-18 for CIFAR10 and CIFAR100, ResNet-34 for ImageNet-1K and RoBERTa for NLP datasets.

\paragraph{Baselines.} (\textit{Supervised}) We compare \algo{} with several score-based and geometry-based coreset selection methods derived from the training dynamics of a model trained on the full dataset as discussed in \citet{zheng2022coverage}: A) \textbf{Random} selection of examples. B) \textbf{Entropy} \citep{colemanselection} of a model's prediction vector. C) \textbf{Forgetting} \citep{tonevaempirical} score for each example i.e., the number of times a model predicts the example incorrectly after having predicted correctly in the previous epoch. D) \textbf{EL2N} \citep{paul2021deep} i.e. L2 norm of error vectors. E) \textbf{Area under the margin} \citep{pleiss2020identifying} score that measures the gap between the prediction probability of the correct target and the next highest probability target. E) \textbf{Moderate} coresets \citep{xia2023moderate} that selects samples at median distance from class center, F) Coverage-based Coreset Selection (\textbf{CCS}) \citep{zheng2022coverage} that divides a range of difficulty scores into equal-sized bins and randomly samples from each bin, and is state-of-art for high pruning rates, G) \textbf{CCS + k-Center}, where k-center samples are selected within each CCS bin, and H) \textbf{BADGE} that selects diverse samples using k-means++ in the gradient vector space. (\textit{Unsupervised}) We compare \algo{} with A) \textbf{Prototypicality} \citep{sorscher2022beyond} that uses self-supervised embeddings to compute k-means clusters and treats samples at a farther distance from the cluster center as more important, B) \textbf{CCS over prototypicality scores}, and C) \textbf{Moderate} coreset selection \citep{xia2023moderate} over the self-supervised embeddings.

\paragraph{Implementation.} In the supervised approach of \algo{}, graph nodes are initialized with supervised difficulty score values and embeddings extracted from the model trained on the entire dataset. We use the forgetting score for CIFAR10, CIFAR100 and AUM score for ImageNet-1K \citep{zheng2022coverage}. We substitute the forgetting score with variance \citep{swayamdipta2020dataset} for NLP datasets since they are trained for fewer epochs and the [CLS] token representation in RoBERTa models for embeddings. Self-supervised \algo{} is initialized with embeddings from SwAV \citep{caron2020unsupervised} for ImageNet-1K and uniform difficulty scores over the dataset.

\paragraph{Computational Complexity of \algo{}.} $k$-nearest neighbors are computed on a A100 GPU using PyTorch, which takes $<$2 minutes for CIFAR10, CIFAR100, Adversarial NLI and ImDB datasets, and approx. 12 minutes for ImageNet-1K. We use \texttt{faiss} indexing (CPU) to get the nearest neighbors for the 12.8 M samples in the Datacomp dataset which takes nearly 55 minutes (8 workers). The iterative selection step in \algo{} has a time complexity of $\mathcal{O}(n)$ in our optimized implementation; for reference, it is completed in $<$5 minutes for DataComp. See details in Appendix.

\paragraph{Algorithm Hyperparameters.} We use the best hyperparameters for baseline methods as reported in the original work. For \algo{}, we set the forward message passing weight $\gamma_{f}$ to 1.0 and perform a sweep over $k =\{1,5,10,15\}$ and $\gamma_{r}=\{0,0.1,0.2...1.0\}$ for CIFAR10, CIFAR100 datasets. Insights from these runs are used to select three configurations for each run on ImageNet-1K; best is reported. See discussion in Sec.~\ref{sec:ablation}.

\section{Results \& Discussion}

\subsection{Comparison to supervised coreset selection methods}

\begin{table}
\caption{\label{tab:vision_results} Results on Vision Datasets. Comparison of performance (acc.) of \algo{} with existing coreset selection methods on CIFAR10, CIFAR100 using ResNet18, and ImageNet-1k using ResNet34 models. Higher is better.}
\setlength{\tabcolsep}{3.1pt}
\centering
\resizebox{\linewidth}{!}{  
\begin{tabular}{lcccccccccccccccccc}
    \toprule
    \textbf{Dataset ($\rightarrow$)} & \multicolumn{6}{c}{\textbf{CIFAR10}} & \multicolumn{6}{c}{\textbf{CIFAR100}} & \multicolumn{6}{c}{\textbf{ImageNet-1K}} \\
    \cmidrule(lr){2-7} \cmidrule(lr){8-13} \cmidrule(lr){14-19}
    \textbf{Pruning Rate ($\rightarrow$)} & \textbf{0\%} & \textbf{30\%} & \textbf{50\%} & \textbf{70\%} & \textbf{80\%} & \textbf{90\%} & \textbf{0\%} & \textbf{30\%} & \textbf{50\%} & \textbf{70\%} & \textbf{80\%} & \textbf{90\%} & \textbf{0\%} & \textbf{30\%} & \textbf{50\%} & \textbf{70\%} & \textbf{80\%} & \textbf{90\%} \\
    \midrule
    \rowcolor{gray!20} \textbf{Random} & 95.5 & 94.3 & 93.4 & 90.9 & 88.0 & 79.0 & 78.7 & 74.6 & 71.1 & 65.3 & 57.4 & 44.8 & 73.1 & 72.2 & 70.3 & 66.7 & 62.5 & 52.3 \\
    \midrule
    \textbf{Entropy} \citep{colemanselection} & - & 94.8 & 92.9 & 90.1 & 84.1 & 72.1 & - & 74.7 & 68.9 & 60.3 & 49.6 & 35.0 & - & 72.3 & 70.8 & 64.0 & 55.8 & 39.0 \\
    \textbf{Forgetting} \citep{tonevaempirical} & - & \textbf{95.7} & 94.9 & 88.1 & 73.8 & 46.3 & - & 76.0 & 68.1 & 49.3 & 30.3 & 20.6 & - & \underline{72.6} & \underline{70.9} & 66.5 & 62.9 & 52.3 \\
    \textbf{EL2N} \citep{paul2021deep} & - & 95.4 & 94.8 & 89.2 & 78.6 & 30.3 & - & 75.6 & 68.1 & 47.2 & 24.8 & 11.8 & - & 72.2 & 67.2 & 48.8 & 31.2 & 12.9 \\
    \textbf{AUM} \citep{pleiss2020identifying} & - & \underline{95.6} & \textbf{95.1} & 87.9 & 68.0 & 40.0 & - & 75.0 & 67.9 & 40.1 & 26.4 & 13.1 & - & 72.5 & 66.6 & 40.4 & 21.1 & 9.9 \\
    \textbf{Moderate} \citep{xia2023moderate} & - & 93.9 & 92.6 & 90.6 & 87.3 & 81.0 & - & 74.6 & 71.1 & 65.3 & 58.5 & 45.5 & - & 72.0 & 70.3 & 65.9 & 61.3 & 52.1 \\
    \textbf{CCS} \citep{zheng2022coverage} & - & 95.4 & \underline{95.0} & \underline{93.0} & 91.0 & \underline{86.9} & - & 77.1 & 74.4 & 68.9 & 64.0 & \textbf{57.3} & - & 72.3 & 70.5 & 67.8 & \underline{64.5} & \textbf{57.3} \\
    \textbf{CCS + k-Center} & - & 95.4 & \textbf{95.1} & 92.9 & \underline{91.1} & 86.8 & - & \underline{77.2} & \underline{74.6} & \underline{69.3} & \underline{64.5} & \underline{57.1} & - & 72.5 & 70.6 & \underline{68.0} & \underline{64.5} & \underline{57.2} \\
     \textbf{BADGE \citep{ash2019deep}} & - & 94.0 & 92.1 & 90.7 & 88.1 & 82.5 & - & 74.7 & 71.8 & 65.2 & 58.9 & 47.8 & - & 71.7 & 70.4 & 65.8 & 61.7 & 53.4 \\
     \midrule
    \textbf{\algo{}} & - & \textbf{95.7} & 94.9 & \textbf{93.3} & \textbf{91.4} & \textbf{87.1} & - & \textbf{78.2} & \textbf{75.9} & \textbf{70.5} & \textbf{65.2} & 56.9 & - & \textbf{72.9} & \textbf{71.8} & \textbf{68.1} & \textbf{65.9} & 55.6 \\
\bottomrule
\end{tabular}
}
\vspace{-10pt}
\end{table}

\begin{table*}[t!]
\caption{\label{tab:nlp_results} Results on NLP Datasets. Comparison of performance (acc.) of \algo{} with existing coreset selection methods on ANLI, ImDB reviews using pretrained RoBERTa$_{\textrm{Large}}$. Higher is better.}
\centering
\resizebox{0.95\linewidth}{!}{  
\begin{tabular}{lcccccccccccc}
\toprule
\textbf{Dataset ($\rightarrow$)} & \multicolumn{6}{c}{\textbf{Adversarial NLI (ANLI)}} & \multicolumn{6}{c}{\textbf{ImDB Reviews (2k)}} \\
\cmidrule(lr){2-7} \cmidrule(lr){8-13}
\textbf{Pruning Rate ($\rightarrow$)} & \textbf{0\%} & \textbf{30\%} & \textbf{50\%} & \textbf{70\%} & \textbf{80\%} & \textbf{90\%} & \textbf{0\%} & \textbf{30\%} & \textbf{50\%} & \textbf{70\%} & \textbf{80\%} & \textbf{90\%} \\
\midrule
    \rowcolor{gray!20} \textbf{Random} & 48.8 & 46.3 & 45.2 & 43.6 & 42.8 & 40.3 & 91.8 & 91.2 & 91.12 & 90.4 & 84.6 & 81.3 \\
    \midrule
    \textbf{Entropy} \citep{colemanselection} & - & \textbf{48.9} & 45.8 & 43.6 & 42.4 & 34.0 & - & 90.6 & 90.4 & 52.8 & 60.1 & 51.3 \\
    \textbf{Variance} \citep{swayamdipta2020dataset} & - & 48.3 & 45.4 & 41.7 & 40.1 & 38.7 & -& 91.4 & 91.0 & 90.2 & 51.5 & 50.7 \\
    \textbf{EL2N} \citep{paul2021deep} & - & 47.7 & \underline{46.3} & 43.9 & 41.1 & 40.3 & - & \underline{91.6} & \underline{91.4} & 51.0 & 50.6 & 50.3 \\
    \textbf{AUM} \citep{pleiss2020identifying} & - & 47.9 & 46.2 & 42.7 & 41.0 & 39.6 & - & \underline{91.6} & \textbf{91.6} & 53.4 & 50.3 & 50.3 \\
    \textbf{Moderate} \citep{xia2023moderate} & - & 46.1 & 44.5 & 43.2 & 42.8 & 40.3 & - & 91.4 & 91.2 & \underline{90.9} & 89.8 & 85.4 \\
    \textbf{CCS} \citep{zheng2022coverage} & - & \underline{48.5} & 46.2 & \underline{44.5} & \underline{43.2} & \textbf{40.4} & - & \underline{91.6} & 90.8 & 90.2 & 89.6 & 87.5 \\
    \textbf{CCS + k-Center} & - & 48.4 & 46.3 & 44.1 & \underline{43.2} & \underline{40.2} & - & 91.4 & 91.0 & 90.6 & \underline{90.2} & 88.2 \\
    \textbf{BADGE} \citep{ash2019deep} & - & 47.3 & 45.8 & 44.0 & 43.1 & 39.5 & - & 91.3 & 90.9 & 90.0 & 90.1 & \underline{89.5} \\
    \midrule
    \textbf{\algo{}} & - & \textbf{48.9} & \textbf{46.7} & \textbf{45.3} & \textbf{44.5} & 40.3 & - & \textbf{91.7} & \textbf{91.6} & \textbf{91.2} & \textbf{90.9} & \textbf{90.3} \\
\bottomrule
\end{tabular}
}
\vspace{-10pt}
\end{table*}

We evaluate \algo{} and other coreset selection methods outlined in Sec.~\ref{sec:setup} on three vision datasets and present results in Tab.~\ref{tab:vision_results}. We observe that \algo{} demonstrates consistent gains over previous state-of-art for all datasets at low and medium pruning rates. \algo{} yields significant gains i.e., 1.0\% and 1.4\%, over the previous best for 50\% and 80\% pruning rates on ImageNet-1K, showing the efficacy of graphs and message passing for coreset selection. Notably, random pruning works surprisingly well for ImageNet-1K, especially for low pruning rates, and is hard to beat. CCS \citep{zheng2022coverage} remains a strong baseline for 90\% pruning rate and only benefits a little from additional diversity-based selection within the CCS bins (see CCS + k-Center in Tab.~\ref{tab:vision_results}). CCS enforces a uniform distribution of sample difficulty scores in the coreset, which is beneficial at high pruning rates for providing even coverage over easy and difficult samples. However, at lower pruning rates (or with increasing data budget), difficult training samples yield a lower test loss from deep learning models \citep{sorscher2022beyond}. The hyperparameters $k$ and $\gamma$ in \algo{} (see Sec.~\ref{sec:our_method}) allow flexibility in the distribution of easy/difficult samples in coresets. We find that higher values of $\gamma$ and lower value of $k$ in \algo{} leads to a coreset that is skewed towards more difficult samples and benefits performance at lower pruning rates. Conversely, low $\gamma$ and high $k$ lead to an equitable distribution over easy/difficult samples and are more useful for higher pruning rates. See discussion on hyperparameters in Sec.\ref{sec:ablation} and qualitative analysis of coresets in Appendix.

Results from the evaluation of various coreset selection methods, including \algo{}, on NLP datasets are presented in Tab.~\ref{tab:nlp_results}. First, we find that when pretrained language models (PLMs) are finetuned on task-specific datasets, the models do not suffer from a catastrophic decline in performance at high pruning rates, in contrast to models trained  from scratch on vision datasets. For IMDB reviews, the performance of finetuned RoBERTa goes from 91.8\% at 0\% pruning to 81.3\% at 90\% pruning using random sampling. The performance improves to 87.5\% using CCS coreset selection and further improves to 90.3\% using \algo{}. The ANLI dataset has been carefully crafted with an iterative, adversarial human-and-model-in-the-loop process, and hence, is significantly less redundant than conventional NLP datasets. The performance for ANLI falls from 48.8\% to 42.8\% at 80\% pruning using random sampling. In this case, CCS coreset selection does not lead to a significant improvement in performance (43.2\%), whereas \algo{} improves the performance by 1.7\% to obtain 44.5\%. Score-based selection methods such as entropy \citep{colemanselection} largely fail to yield results better than random pruning at high pruning rates. These results show that \algo{} is effective for both language and vision modalities.

\begin{figure}
    \vspace{-15pt}
    \centering
    \includegraphics[width=0.96\linewidth]{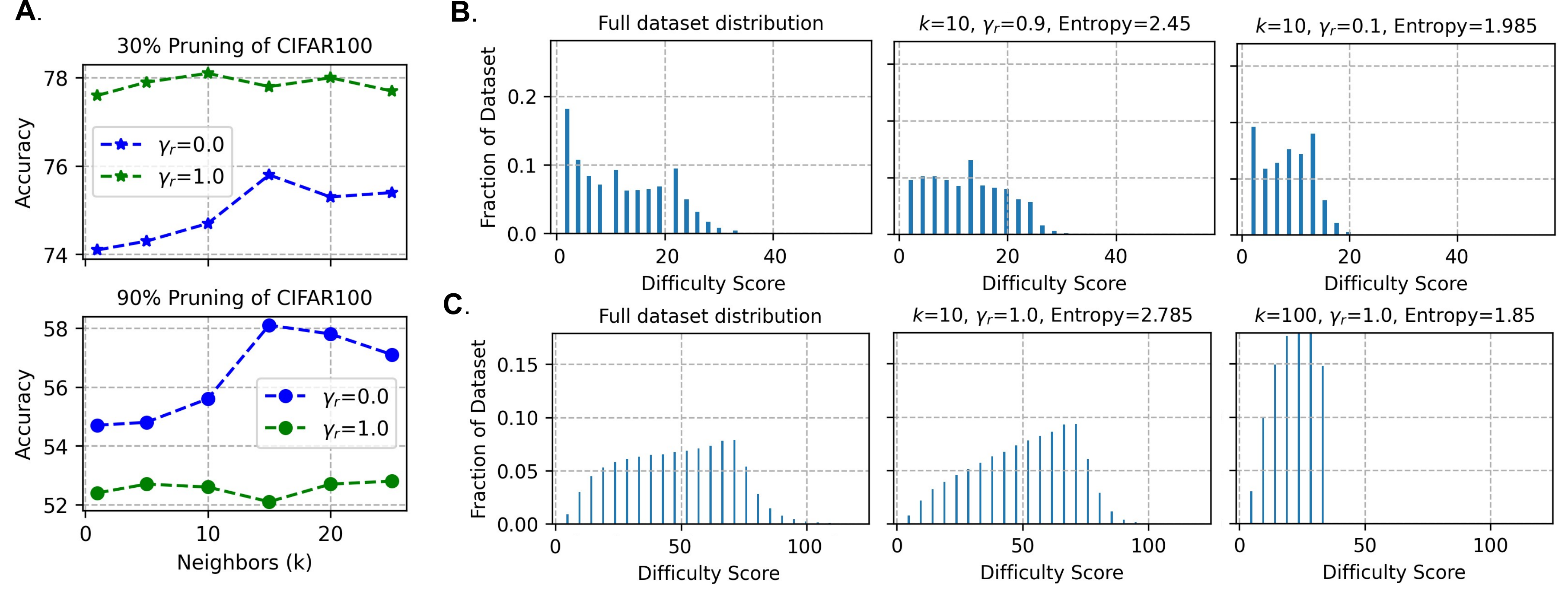}
    \vspace{-10pt}
    \caption{Effect of $k$, $\gamma_{r}$. (A) Accuracy at 30\%, 90\% pruning of CIFAR100 for nearest neighbors ($k$) and message passing weight $\gamma_{r}$ values; Distribution of difficulty scores in the best coresets selected via \algo{} for 30\% (center) and 70\% (right) pruning of (B) CIFAR100, (C) ImageNet-1K.}
    \label{fig:ablation_k}
    \vspace{-10pt}
\end{figure}

\subsection{Analysis of \algo{}}
\label{sec:ablation}
\algo{} contains two hyperparameters, $k$ nearest neighbors and reverse message passing weight $\gamma_{r}$ (see Sec.~\ref{sec:our_method}) that allow various distributions of importance scores in the selected coreset. We conduct experiments to analyze their effect on CIFAR100 performance and present results in Fig.~\ref{fig:ablation_k}.

At low pruning rates (see top, Fig.~\ref{fig:ablation_k}(a)), higher $k$ has a small effect on performance when the updates during reverse message passing are weak ($\gamma_{r}$=1.0). However, the coresets selected at high $k$ and low $\gamma_{r}$ include a majority of the difficult samples from the full dataset, which works best for low pruning rates on CIFAR100, as demonstrated by the distribution of importance scores in best-performing coreset at 30\% pruning rate (see Fig.~\ref{fig:ablation_k}(B), center). We use this insight to pick a similar configuration of \algo{} for ImageNet-1K and find that it transfers well. The distribution of difficulty scores in the best-performing coreset of ImageNet-1K at 30\% pruning rate is presented in Fig.~\ref{fig:ablation_k}(C).

Higher $k$ improves performance when large updates ($\gamma_{r}$=0.0) are being made to the nodes connected to the selected node at high pruning rates (see bottom, Fig.~\ref{fig:ablation_k}(a)). This is because low $\gamma_{r}$ value leads to aggressive downweighting of semantically similar samples when a sample is selected and promotes diversity under a fixed data budget. The selected samples also form an equitable distribution over a small range of difficulty scores. Consequently, such coresets work best for medium-to-high pruning rates, as evidenced by the distribution of difficulty scores in the best performing coresets at 70\% pruning rate for CIFAR100 and ImageNet-1K (see Fig.~\ref{fig:ablation_k}(B,C), right).

\subsection{Self-supervised and unsupervised approaches using \algo{}}

Existing methods for obtaining sample difficulty scores and coresets generally rely on a model trained on the full dataset, which undermines their utility for curating new datasets. Hence, we adopt \algo{} for self-supervised and unsupervised data selection approaches, and show promising results that motivate further research in this direction.

\paragraph{Unsupervised data filtering.} \citet{gadre2023datacomp} show that a simple strategy of retaining the samples with a high CLIP score is a strong baseline filtering method (see Tab.~\ref{tab:datacomp}) on DataComp, a massive unfiltered corpus of images and texts to train CLIP-style models \citep{radford2021learning}.\footnote{Our reproduced numbers are lower than \citet{gadre2023datacomp} because some images in the original corpus fail download. We report improvements using \algo{} on this subset of images for fair comparison.} However, a strategy based on individual sample scores only ignores potential redundancies in the dataset and may allot unnecessary data budget to an easy but dense region of the sample space. Hence, we adapt \algo{} for filtering DataComp by treating the CLIP score as the difficulty score and using CLIP embeddings for computing sample distances. Results are presented in Tab.~\ref{tab:datacomp}. We find that the data selected by \algo{} using both, CLIP text and image embeddings, for computing sample distances improves average zero-shot performance on 38 image classification and multimodal datasets by 1\% at the same data budget. Notably, it improves performance on the diverse set of VTAB image classification datasets \citep{zhai2019large} by nearly 4\%, demonstrating the importance of diverse training datasets for learning generalizable representations. When the similarity is computed using only text embeddings, we see smaller improvements in retrieval tasks and average performance i.e., 0.4\% and 0.6\% respectively. The retrieval performance is highest using the CCS strategy \citep{zheng2022coverage} which samples from the entire range of CLIP scores, however, it significantly hurts performance on other tasks. Computing distances between samples using image embeddings only in \algo{} does not improve the average performance and hurts performance on ImageNet-1K.

\paragraph{Self-supervised coreset selection.}  \citet{sorscher2022beyond} use embeddings from SwAV \citep{caron2020unsupervised}, a model trained on ImageNet-1k in a self-supervised manner, and use the spatial distribution of the samples in the embedding space to assign difficulty scores (prototypicality).  We adopt 

\begin{wrapfigure}{r}{0.45\textwidth}
\vspace{-20pt}
\centering
    \includegraphics[width=0.43\textwidth]{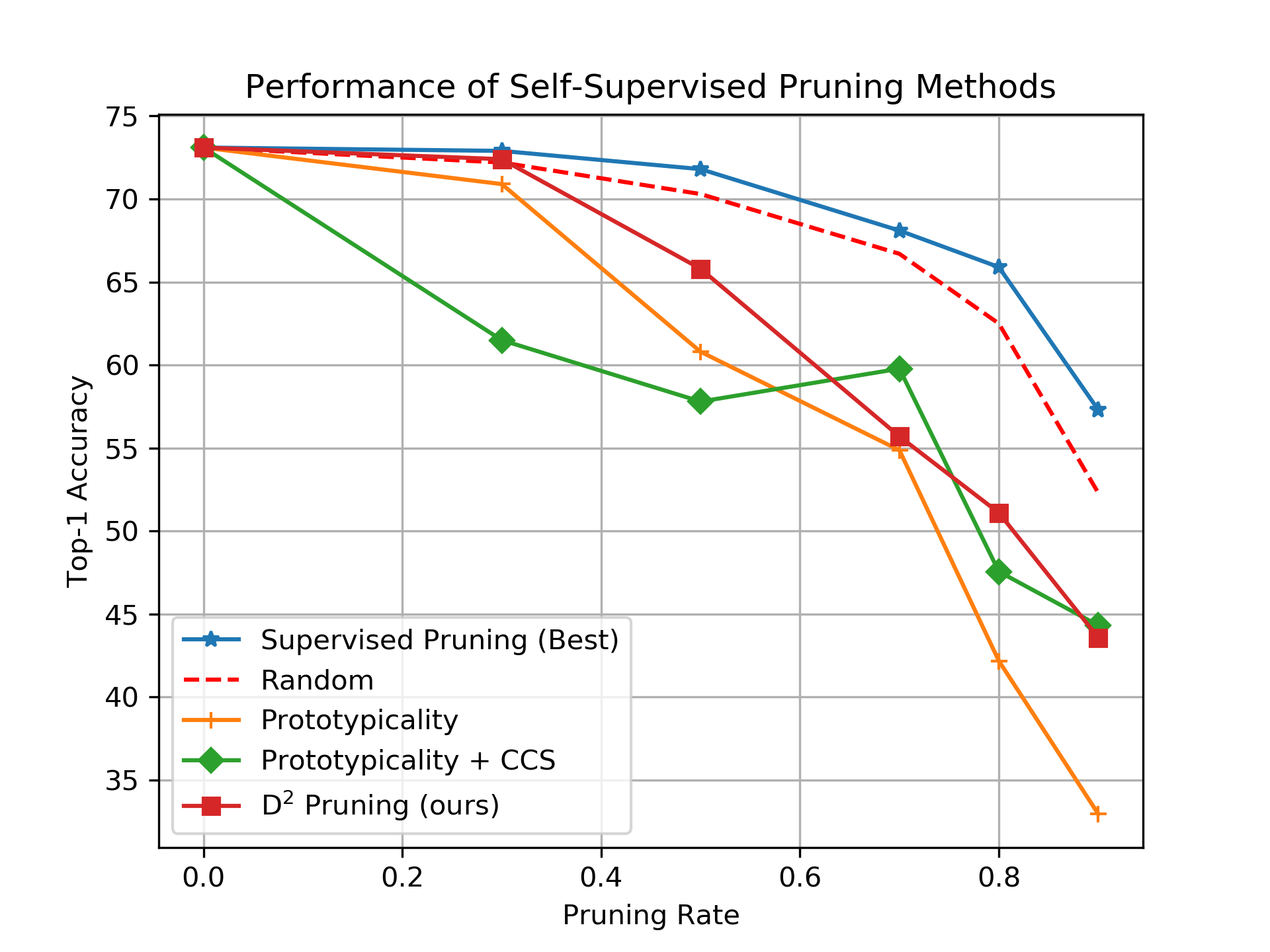}
  \caption{Results of self-supervised pruning methods on ImageNet-1K. \algo{} performs as good as the best supervised pruning method at 30\% pruning rate and significantly improves over other self-supervised methods.\label{fig:self}}
  \vspace{-10pt}
\end{wrapfigure}

\noindent \algo{} for a similar self-supervised approach by using SwAV embeddings to compute sample distances and initialize node features with a unit value. In the absence of difficulty scores, \algo{} ranks the samples solely by the density of their neighborhood in the embedding space. See results in Fig. ~\ref{fig:self}. Prototypicality suffers drastically at over 30\% pruning rates. When combined with CCS, it yields 10\% gain for 90\% pruning rate and lesser gains for 70\%, 80\% pruning rates. \algo{} further improves performance by 3\% at 80\% pruning rate and provides similar gains over prototypicality for lower pruning rates i.e., 1\%, 5\% at 30\% and 50\% pruning rates respectively. These self-supervised pruning methods fall short of performance from random pruning; nevertheless, results on \algo{} demonstrate that better ways of manipulating the spatial structure of datasets are useful.

\begin{table*}[t!]
\centering
\caption{\label{tab:datacomp} Results on DataComp. Comparison of performance (acc.) of \algo{} with CCS \citep{zheng2022coverage} and data filtering methods presented in \citet{gadre2023datacomp}. Higher is better.}
\small
\resizebox{0.95\linewidth}{!}{ 
\begin{tabular}{lcccccc}
    \toprule
    \textbf{Filtering Strategy} & \textbf{Dataset Size} & \textbf{ImageNet} & \textbf{ImageNet Dist. Shift} & \textbf{VTAB} & \textbf{Retrieval} & \textbf{Average} \\
    \midrule
    \rowcolor{gray!20} No filtering \citep{gadre2023datacomp} & 12.8M & 2.5 & 3.3 & 14.5 & 11.4 & 13.2 \\
    Text-based filtering \citep{gadre2023datacomp} & 3.2M & 4.6 & 5.2 & 16.9 & 12.5 & 15.7 \\
    Image-based filtering \citep{gadre2023datacomp} & 3.2M & 4.3 & 4.7 & 17.8 & 12.1 & 15.9 \\
    CLIP score (L/14 30\%) \citep{gadre2023datacomp} & 3.8M & 5.1 & 5.5 & 19.0 & 11.7  & 17.3 \\
    \midrule
    CLIP score (L/14 30\%, reproduced) & 3.8M & \textbf{5.1} & \textbf{5.6} & 17.0 & 11.9 & 16.0 \\
    \midrule
    CCS \citep{zheng2022coverage} & 3.8M & 2.6 & 3.7 & 14.3 & \textbf{14.2} & 13.8 \\
    \textbf{\algo{}} (image + text) & 3.8M & \textbf{5.1} & \textbf{5.6} & \textbf{18.2} & 11.7 & \textbf{17.0} \\
    \textbf{\algo{}} (image only) & 3.8M & 4.4 & 5.1 & 16.9 & 12.1 & 15.9 \\
    \textbf{\algo{}} (text only) & 3.8M & \underline{4.9} & \underline{5.5} & \underline{17.0} & \underline{12.3} & \underline{16.6}\\
    \bottomrule
\end{tabular}
}
\vspace{-10pt}
\end{table*}

\section{Analysis \& Discussion}
\begin{figure}
    \centering
    \includegraphics[width=0.93\linewidth]{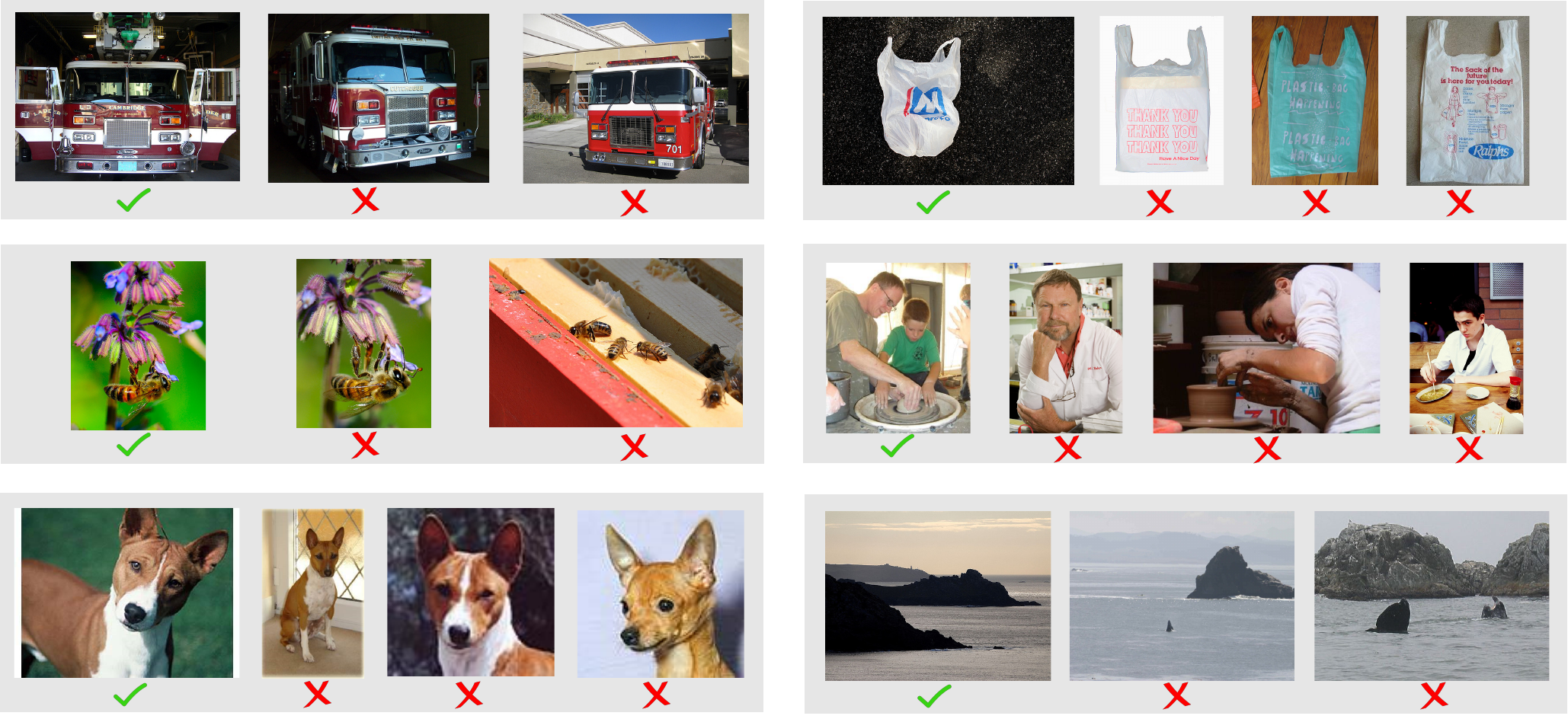}
    \vspace{-10pt}
    \caption{Example of coresets selected by \algo{} from ImageNet-1K at 30\% pruning rate. Image sub-populations are extracted from ImageNet-1K by a recursive traversal of the connectivity graph $\mathcal{G}$ initialized for \algo{}. For each sub-population, we show the images retained in the coreset with \textcolor{green}{\checkmark} and the images left out of the coreset with \textcolor{red}{\text{\sffamily X}}.}
    \label{fig:samples}
\end{figure}

\paragraph{Qualitative analysis of coresets selected by \algo{}.} In order to perform a qualitative analysis of the merits of \algo{}, we first use the connectivity graph $\mathcal{G}$ to extract meaningful sub-populations from the entire ImageNet-1K dataset. For each sample, we recursively seek nearest neighbors that are situated at a distance in the embedding space that is less than a predefined threshold. Next, for each of these sub-populations, we differentiate the samples that appear in the coreset selected by \algo{} at 30\% pruning of ImageNet-1K. We present and analyze a few representative sub-populations in Fig.~\ref{fig:samples}. First, we observe several cases where \algo{} successfully avoids selecting perceptual duplicates \citep{abbas2023semdedup} in the coreset (see top left and middle left in Fig.~\ref{fig:samples}). Next, we see multiple cases where a composite image is selected for the coreset, and images that contain one or more of the subjects/objects in the selected image are left out (see middle right in Fig.~\ref{fig:samples}). Finally, we find that relying on the semantic similarity of pretrained embeddings can lead to the propagation of errors, as seen in the sub-population on the bottom right in Fig.~\ref{fig:samples}. The images that contain dolphins are left out of the coreset because of their similarity to an image depicting a water landscape.

\paragraph{Visualization of data distribution in coresets.} We showcase the results of various sampling methods for a single class in the CIFAR10 dataset in Fig.~\ref{fig:sampling}. The embeddings are obtained from a ResNet18 network trained on the full training dataset and compressed to two dimensions using PCA (~90\% explained variance) for simpler visualization. As seen in Fig.~\ref{fig:sampling}(b), random sampling leads to relatively larger samples from the denser region of the distribution and consequently, a higher percentage of easy samples feature in the coreset after 90\% pruning. By optimizing for diversity only via greedy $k$-center selection (Fig.~\ref{fig:sampling}(c)), the diversity of the coreset remains high but it is plagued with the same problem as random sampling i.e. easier samples are preferred. Moderate coresets \citep{xia2023moderate} sample from a narrow area in the distribution, resulting in poor diversity and a slightly better balance between easy and difficult samples (Fig.~\ref{fig:sampling}(d)). Finally, with our proposed method, the diversity remains high and the distribution of difficulty scores in the coreset is also balanced (Fig.~\ref{fig:sampling}(f)). In Fig.~\ref{fig:sampling}(e), we assign unit values to node features instead of the corresponding difficulty scores (see Sec.~\ref{sec:our_method}); the resulting coreset is not much different from random sampling, showing that our proposed approach is crucial for balancing difficulty and diversity in coresets.

\section{Related Work}
\label{related_work}

\textbf{Coreset Selection.} Coreset selection has been widely studied in machine learning \citep{welling2009herding, chen2010super, feldman2011scalable}. Recent works have focused on large datasets and deep networks. Geometry-based methods remove redundant information \citep{welling2009herding, seneractive, pooladzandi2022adaptive}. Uncertainty/loss/error-based methods estimate the difficulty of a sample from model confidence \citep{swayamdipta2020dataset} or its training dynamics \cite{tonevaempirical, paul2021deep, bachem2015coresets}. Submodular functions \citep{wei2015submodularity}, gradient-matching \citep{mirzasoleiman2020coresets}, and optimization \citep{yang2022dataset, tukan2023provable} have been explored for coreset selection. Relevant works are also termed \textit{data distillation} \citep{cazenavette2022dataset} or \textit{data pruning} \citep{sorscher2022beyond}. We combine data diversity and sample difficulty into a unified coreset selection algorithm.

\noindent \textbf{Data Pruning in NLP}. Works exploring coreset selection methods for NLP datasets have been far and few \citep{fayyazbert}. \citet{abbas2023semdedup} removes semantic duplicates from C4 dataset \citep{raffel2020exploring} to reduce data size and improve performance. \citet{kaddour2023minipile} introduce a small version of the Pile dataset \citep{gao2020pile} for pretraining BERT \citep{devlin2018bert, liu2019roberta}. We evaluate coreset selection methods on sentiment analysis, natural language inference tasks.

\noindent \textbf{Message Passing for Coreset Selection.}  Neural message passing \citep{yadav19lcn,yadati2019hypergcn} is well-explored in graph neural networks for chemical structures \citep{gilmer2017neural}, however, has seen less exploration in the representation of datasets. \citet{ebert2012ralf} use a message-passing framework based on embedding-distance only for performing graph-based density sampling during active learning. \citet{kim2021message} use message-passing to learn the topology of input data in online learning.

\section{Conclusion}
We introduce a novel coreset selection algorithm, \algo{}, based on message-passing within a graph representing the dataset. Our algorithm combines data diversity and difficulty to select a coreset that outperforms existing coreset selection methods at low-to-medium pruning rates on multiple vision and NLP benchmarks, and can be adapted into self-supervised, unsupervised data selection.

\paragraph{Acknowledgement.} This work was supported by ARO Award W911NF2110220, ONR Grant N00014-23-1-2356, and NSF-AI Engage Institute DRL-211263, and DARPA MCS Grant N66001-19-2-4031. The views, opinions, and/or findings contained in this article are those of the authors and not of the funding agency.

\bibliography{iclr2024_conference}
\bibliographystyle{iclr2024_conference}

\appendix
\section*{Overview}
\noindent The appendix is organized as follows:\\
\textbf{Section A}: Code release.\\
\textbf{Section B}: Details of the datasets and the best hyperparameters for our models.\\
\textbf{Section C}: Computational complexity of \algo{} for all datasets.\\
\textbf{Section D}: Limitations and license.

\section{Code Release}
Code for all experimental results reported in our paper is available with the supplementary submission.

\section{Datasets \& Hyperparameters}
\subsection{Datasets}
\paragraph{Vision Benchmarks.} We use the CIFAR10, CIFAR100 \citep{krizhevsky2009learning} and ImageNet-1K \citep{deng2009imagenet} image classification datasets for our experiments on vision benchmarks. The CIFAR10 dataset consists of 60000 32x32 color images for 10 classes, with 6000 images per class. The training and test splits contain 50000 and 10000 images respectively. The CIFAR100 dataset has 100 classes containing 500 and 100 images per class in the training and test splits respectively. Details about the class labels in CIFAR10, CIFAR100 datasets can be found \href{https://www.cs.toronto.edu/~kriz/cifar.html}{here}. The ImageNet-1K dataset comprises approximately 1.2 million real-world images distributed over 1000 object classes. It contains 1,281,167 and 50,000 images in training and validation splits respectively.

\paragraph{NLP Benchmarks.} We select two popularly used NLP tasks i.e. natural language inference (NLI) \citep{bowman2015large} and sentiment analysis \citep{turney2002thumbs}. For natural language inference, we use the Adversarial NLI dataset \citep{nie2020adversarial} that has been created in an iterative human-and-model-in-the-loop adversarial procedure. During each iteration, human annotators are instructed to devise examples that the current best models are unable to answer correctly. The models are trained on these challenging annotations for stronger performance. Multiple rounds of such iterations result in a challenging NLI benchmark. We use the data created in the third (and final) round of this process which contains 100459, 1200, and 1200 examples in the training, development, and test splits respectively. We use the ImDB reviews dataset \citep{maas2011learning} for the sentiment analysis task. The original dataset contains 25000 examples each in the training and test splits and is a binary classification dataset. Our experiments showed that models trained on 10\% of this dataset achieved nearly the same performance as 100\% of the dataset. We observed similar trends for other popular sentiment analysis benchmarks as well such as Yelp Reviews \citep{zhang2015character}, SST2 \citep{socher2013recursive} etc. Hence, we created an in-house version of the ImDB Reviews dataset that contains 2000, and 1000 samples in the training and development splits respectively, that are randomly selected from the original training set. We retain the original test split containing 25000 samples for evaluation in our experiments.

\subsection{Training Hyperparameters}

\paragraph{Coreset Selection.} We use the recommended hyperparameters in \citet{zheng2022coverage} for experiments using Coverage-based coreset selection (CCS) i.e. 50 bins (or \textit{strata}) for all pruning rates. Models trained on vision datasets are also subjected to a hard cutoff rate $\beta$ on the difficulty score for eliminating outliers or erroneous samples (see \citet{zheng2022coverage} for the values). We report the best hyperparameters for \algo{} in Tabs.~\ref{tab:coreset_params}\&~\ref{tab:coreset_params_2}.

\begin{table}
\caption{\label{tab:coreset_params} Best values of nearest-neighbors ($k$) and reverse message passing weight ($\gamma_{r}$) for vision datasets. See a discussion on these hyperparameters in Sec.~\ref{sec:ablation}.}
\setlength{\tabcolsep}{3.1pt}
\centering
\resizebox{\linewidth}{!}{  
\begin{tabular}{lcccccccccccccccccc}
    \toprule
    \textbf{Dataset ($\rightarrow$)} & \multicolumn{6}{c}{\textbf{CIFAR10}} & \multicolumn{6}{c}{\textbf{CIFAR100}} & \multicolumn{6}{c}{\textbf{ImageNet-1K}} \\
    \cmidrule(lr){2-7} \cmidrule(lr){8-13} \cmidrule(lr){14-19}
    \textbf{Pruning Rate ($\rightarrow$)} & \textbf{0\%} & \textbf{30\%} & \textbf{50\%} & \textbf{70\%} & \textbf{80\%} & \textbf{90\%} & \textbf{0\%} & \textbf{30\%} & \textbf{50\%} & \textbf{70\%} & \textbf{80\%} & \textbf{90\%} & \textbf{0\%} & \textbf{30\%} & \textbf{50\%} & \textbf{70\%} & \textbf{80\%} & \textbf{90\%} \\
    \midrule
    \textbf{Nearest Neighbors ($k$)} & - & 10 & 5 & 1 & 2 & 2 & - & 10 & 10 & 10 & 5 & 15 & - & 50 & 50 & 100 & 10 & 10 \\
    \midrule
    \textbf{Reverse Message Passing ($\gamma_{r}$)} & - & 0.9 & 1.0 & 0.1 & 0.0 & 0.0 & - & 0.9 & 0.8 & 0.3 & 0.3 & 0.0 & - & 1.0 & 1.0 & 0.3 & 0.1 & 0.0 \\
\bottomrule
\end{tabular}
}
\end{table}

\begin{table}
\caption{\label{tab:coreset_params_2} Best values of nearest-neighbors ($k$) and reverse message passing weight ($\gamma_{r}$) for NLP datasets and self-supervised \algo{} of ImageNet-1K. See details in Sec.~\ref{sec:ablation}.}
\setlength{\tabcolsep}{3.1pt}
\centering
\resizebox{\linewidth}{!}{  
\begin{tabular}{lcccccccccccccccccc}
    \toprule
    \textbf{Dataset ($\rightarrow$)} & \multicolumn{6}{c}{\textbf{Adversarial NLI}} & \multicolumn{6}{c}{\textbf{ImDB(2K)}} & \multicolumn{6}{c}{\textbf{ImageNet-1K (self-supervised)}} \\
    \cmidrule(lr){2-7} \cmidrule(lr){8-13} \cmidrule(lr){14-19}
    \textbf{Pruning Rate ($\rightarrow$)} & \textbf{0\%} & \textbf{30\%} & \textbf{50\%} & \textbf{70\%} & \textbf{80\%} & \textbf{90\%} & \textbf{0\%} & \textbf{30\%} & \textbf{50\%} & \textbf{70\%} & \textbf{80\%} & \textbf{90\%} & \textbf{0\%} & \textbf{30\%} & \textbf{50\%} & \textbf{70\%} & \textbf{80\%} & \textbf{90\%} \\
    \midrule
    \textbf{Nearest Neighbors ($k$)} & - & 15 & 10 & 5 & 5 & 5 & - & 10 & 10 & 10 & 5 & 2 & - & 50 & 100 & 25 & 10 & 25 \\
    \midrule
    \textbf{Reverse Message Passing ($\gamma_{r}$)} & - & 1.0 & 1.0 & 0.1 & 0.1 & 0.0 & - & 1.0 & 0.8 & 0.3 & 0.0 & 0.0 & - & 1.0 & 1.0 & 0.5 & 0.5 & 0.0 \\
\bottomrule
\end{tabular}
}
\end{table}

\paragraph{Models.} We follow the best training hyperparameters for ResNet18 model and ResNet34 models as suggested in \citet{zheng2022coverage} to remain comparable to the numbers reported in their work. For fine-tuning of pretrained RoBERTa on NLP datasets, we perform a grid search over learning rates $\{1e^{-5}, 2e^{-5}, 5e^{-5}, 1e^{-4}\}$ and batch sizes $\{8, 16, 32\}$ using 100\% of the data, which results in learning rate of $1e^{-4}$ and batch size of 32 for Adversarial NLI, ImDB (2k) datasets. Models are trained on pruned datasets using the same hyperparameters that are used for training 100\% of the data. The maximum number of training steps is kept constant across all pruning rates. RoBERTa models are trained for 10000 and 1500 training steps for Adversarial NLI and ImDB (2k) datasets respectively, with early stopping.

\begin{table*}
\caption{\label{tab:time_results} Computational Overhead for \algo{}. Comparison of runtime of \algo{} for 100\% selection of the various datasets in our experiments. \algo{} can be divided into the `Graph creation' and `Iterative selection' steps (see General Response). Larger datasets like DataComp have a `\texttt{faiss} indexing' step to enable fast nearest-neighbor lookup. Results are computed using a multi-thread implementation of \algo{} using 8 workers on a CPU with 32 cores.}
\setlength{\tabcolsep}{3.1pt}
\centering
\resizebox{0.9\linewidth}{!}{  
\begin{tabular}{lcccccc}
    \toprule
    \textbf{Dataset ($\rightarrow$)} & \textbf{CIFAR10} & \textbf{CIFAR100} & \textbf{Adv. NLI} & \textbf{ImDB} & \textbf{DataComp} & \textbf{ImageNet-1K}\\
    \midrule
    \textbf{\texttt{faiss} indexing} & - & - & -  & - & 25m & - \\
    \textbf{Graph creation} & 2m & 1m & 4m & 1m & 30m & 15m \\
    \textbf{Iterative selection} & 1m & 1m & 2m & 1m & 7m & 8m \\
    \midrule
    \textbf{Total Time} & 3m & 2m & 6m & 2m & 1h 2m & 23m \\
\bottomrule
\end{tabular}
}
\end{table*}

\section{Computational complexity of \algo{}}

We divide the runtime into 1. ‘Graph creation’ which includes graph initialization and forward message passing, and 2. ‘Iterative selection’ (see Sec.~\ref{sec:our_method}) and present results in Tab.~\ref{tab:time_results} for 100\% data selection of the various datasets used in our experiments. Numbers are rounded to the nearest minute. Runtime for iterative selection is proportional to the size of the coreset being selected. Hence, in practice, the runtime for iterative selection is even lower since we only select a subset of the data in our experiments.

\section{Limitations \& License}
\subsection{Limitations}
\paragraph{Access to Full Dataset \& Pretrained Model.} Similar to the many previous coreset selection methods, our method relies on a model that has been pretrained or finetuned on the full dataset. We leverage the pretrained embeddings as well as the difficulty scores from this model. In doing so, we risk capturing the biases of the model. Further, one cannot use \algo{} to create datasets from scratch and reduce annotation costs by avoiding redundant samples in the dataset. We note that an ideal data pruning method would not rely on access to the full dataset so that it can be used for creating challenging and effective datasets in a cost-effective manner. Our experiments in self-supervised and unsupervised data selection show promising results in this direction.

\subsection{License}
We will publicly release our code and models. We use standard licenses from the community and provide the following links to the licenses for the datasets that we used in the
project.\\

\textbf{CIFAR10, CIFAR100:} \href{https://www.cs.toronto.edu/~kriz/cifar.html}{Other}\\
\textbf{Adversarial NLI:} \href{https://github.com/facebookresearch/anli/blob/main/LICENSE}{Creative Commons}\\
\textbf{ImDB Reviews:} \href{http://ai.stanford.edu/~amaas/data/sentiment/}{Other}\\
\textbf{Counterfactual ImDB, NLI:} \href{https://github.com/acmi-lab/counterfactually-augmented-data/blob/master/LICENSE}{Apache}\\
\textbf{DataComp:} \href{https://github.com/mlfoundations/datacomp/blob/main/LICENSE}{MIT}

\end{document}